\documentclass{article}

\PassOptionsToPackage{numbers, sort&compress}{natbib}

 \usepackage[dblblindworkshop, final]{neurips_2026}
\workshoptitle{AI4Mat}

\usepackage[utf8]{inputenc} 
\usepackage[T1]{fontenc}    
\usepackage[hidelinks]{hyperref}       
\usepackage{url}            
\usepackage{booktabs}       
\usepackage{amsfonts}       
\usepackage{nicefrac}       
\usepackage{microtype}      
\usepackage{xcolor}         

\usepackage{graphicx}       
\usepackage[version=4]{mhchem}

\title{Does System Size Matter in Uncertainty Quantification for Machine Learning Interatomic Potentials?}

\author{%
  David S.~Hippocampus\thanks{Use footnote for providing further information
    about author (webpage, alternative address)---\emph{not} for acknowledging
    funding agencies.} \\
  Department of Computer Science\\
  Cranberry-Lemon University\\
  Pittsburgh, PA 15213 \\
  \texttt{hippo@cs.cranberry-lemon.edu} \\
}

\author{%
  Chayaphol Lortaraprasert\thanks{Work done during an internship at Preferred Networks. Now at the National University of Singapore: \texttt{chayaphol.l@u.nus.edu} } \\
  Massachusetts Institute of Technology \\
  \And
  Nontawat Charoenphakdee \\
  Preferred Networks \\
  \texttt{nontawat@preferred.jp} \\
  \And
  Wenwen Li \\
  Preferred Networks \\
  \texttt{wenwenli@preferred.jp} \\
  \And
  Yong-Bin Zhuang \\
  Preferred Networks \\
  \texttt{ybzhuang@preferred.jp} \\
  \And
  Yuta Tsuboi \\
  Preferred Networks \\
  \texttt{tsuboi@preferred.jp} \\
}

\begin{document}

\maketitle

\vspace{-2mm}
\begin{abstract}
Machine-learning interatomic potentials (MLIPs) commonly use uncertainty estimates to flag out-of-distribution (OOD) configurations and select structures for active learning. 
A key question is whether the resulting structure-level uncertainty scores remain comparable across different system sizes. 
A common practice is to aggregate per-atom uncertainties by their mean or maximum and compare the resulting score against a fixed threshold, implicitly assuming such comparability. 
We test this assumption across several in-distribution (ID)/OOD pairs by characterizing both aggregations as a function of atom count. 
The two aggregations exhibit distinct size dependencies. 
Mean aggregation remains size-independent for homogeneous systems but loses sensitivity to defects in large cells. 
In contrast, maximum aggregation grows with system size for ID systems while remaining approximately flat for defect-containing systems, narrowing the ID--OOD gap in a manner consistent with extreme-value statistics. 
Although highly uncertain atoms in large structures may indicate novel local environments, retraining on these structures yields little improvement, in contrast to retraining on OOD structures constructed to lie outside the training distribution.
Our results reveal a practical limitation of size-unaware uncertainty criteria and motivate size-aware calibration for more reliable OOD detection in large-scale atomistic simulations.
\end{abstract}
\vspace{-2mm}

\vspace{-2mm}
\section{Introduction}
\label{sec:intro}
\vspace{-2mm}

Machine learning interatomic potentials (MLIPs) accelerate atomistic simulations for materials discovery by fitting the potential energy surface from first-principles reference data and evaluating it orders of magnitude faster than density functional theory (DFT), giving near-DFT accuracy at a cost that scales linearly with the number of atoms
\citep{Behler2007,Bartok2010,Behler2016}. 
Successive architectures, e.g., descriptor-based neural networks \citep{Behler2007,Smith2017}, kernel models \citep{Bartok2010}, graph neural networks \citep{Schutt2018}, and equivariant message-passing
models \citep{Batzner2022,Batatia2022} have steadily improved accuracy and data efficiency. 
More recently, universal or foundation potentials trained on large DFT databases \citep{Chen2022,Deng2023,Batatia2025,Merchant2023,Takamoto2022,Rhodes2025} have made it routine to run atomistic simulations across the periodic table without system-specific fitting, and MLIP-driven molecular dynamics now reaches system sizes and time scales far beyond what DFT allows \citep{kozinsky2023scaling,Jia2020}.

MLIPs accuracy typically  degrades on local atomic environments absent from the training set, and this degradation is silent without an uncertainty estimate \citep{Behler2016,Kulichenko2021}.
Uncertainty quantification (UQ) can be useful to flag unreliable predictions during simulation~\citep{Schran2020,Imbalzano2021,Vandermause2020}.
Moreover, it can drive active learning loops that select structures for DFT labeling and iterative retraining \citep{Podryabinkin2017,Smith2018,Zhang2019,Vandermause2020,Kulichenko2023,Schwalbe2021}.
Common UQ metrics include ensemble disagreement \citep{Lakshminarayanan2017,Musil2019,Schran2020}, Gaussian process~\citep{Vandermause2020}, distance or committee-free measures based on descriptor-space novelty \citep{Zhu2023,Janet2019}.

Growing compute has pushed MLIP-driven simulations towards ever larger
systems of up to millions of atoms in production MD~\citep{Jia2020,kozinsky2023scaling}, and active learning datasets become increasingly mixed-size.
Structure-level uncertainty scores, however, are built without size consideration.
Per-atom uncertainties, including force uncertainties~\citep{Zhang2019,Vandermause2020,jinnouchi2019phase,Schran2020}
and descriptor-space extrapolation grades~\citep{Podryabinkin2017,novikov2021mlip,huang2025p}, are reduced to a single
score by the maximum ~\citep{Zhang2019,zhang2020dp,Podryabinkin2017,novikov2021mlip,Vandermause2020,jinnouchi2019phase,zaverkin2024uncertainty},
or mean of the atom-wise score over all atoms~\citep{Schran2020,Schwalbe2021,tan2023single}, a choice
presented as convention rather than justified; other workflows threshold
a global energy disagreement
directly~\citep{Smith2018,Kulichenko2023}. 
The score is then compared against a fixed threshold calibrated on small reference cells~\citep{zhang2020dp,novikov2021mlip,Vandermause2020,zaverkin2024uncertainty}.
Both conventions thus assume that a structure-level score is comparable across system sizes, which could affect UQ scores.
However, previous studies on system size effects are only limited to total energy uncertainty~\citep{Smith2018,kellner2024uncertainty}. 
Whether the maximum or mean UQ stays comparable as the system size grows is still an open question.

In this work, we investigate how mean and maximum per-atom force uncertainty scale with system size for multiple different in-distribution (ID)/OOD pairs using MACE~\citep{Batatia2022} and ORB~\citep{Rhodes2025} ensembles and a single-model Force-Delta uncertainty estimate.
We find that mean uncertainty is size-independent for homogeneous systems and converges to pristine for defective ones,
while maximum uncertainty increases with system size for ID systems and stays flat for a defect, eliminating the ID/OOD gap at large sizes and hence could fail OOD detection based on size-unaware UQ score.
We attribute this to extreme-value statistics and verify that the atoms with max uncertainty in large structures have unfamiliar local environment and high error,
yet find that retraining on these structures UQ flags as OOD yields little model improvement, whereas retraining on small OOD-by-design structures is cheaper and improves the model significantly.
This shows a limitation of current UQ methods that do not take size into account for OOD detection.
In Appendix~\ref{app:calibrator}, we explore a trainable post-hoc, size-aware uncertainty calibrator and find that removing size bias consistently across systems is nontrivial.
Overall, system size biases UQ in MLIPs and can cause OOD detection to fail. 
These results pave the way for size-aware post-hoc calibration, or training losses that account for outlier atoms, toward more reliable OOD detection in large-scale MLIP-accelerated materials simulations and discovery.
\vspace{-1mm}
\section{Results and Discussion}
\label{sec:results}

\vspace{-1mm}
\subsection{Uncertainty Scaling with System Size}
\label{sec:scaling}
We use force uncertainties $u_i$ for atom $i$ and summarize a structure of $N$ atoms by the two aggregation methods used for structure selection: the mean $\bar{u} = N^{-1}\sum_i u_i$ and the maximum $u_{\max} = \max_i u_i$.
We also investigate percentile-based aggregation methods in Appendix \ref{app:pc-uq}.
Our main uncertainty metric ($u_i$)  is the ensemble variance of predicted forces across independently initialized models \citep{Lakshminarayanan2017,Musil2019,Schran2020,Imbalzano2021},
computed for MACE \citep{Batatia2022} and ORB \citep{Rhodes2025}.
We also investigate the ORB Force Delta, defined as the difference between non-conservative and conservative forces \citep{Lortaraprasert2025}, as a single-model UQ alternative which gives the same results as ensemble UQs in Appendix \ref{app:force-delta}.
We study three different ID/OOD pairs: pristine versus vacancy-containing \ce{Ni3Al} (mp-2593 from the Materials Project dataset \citep{Jain2013}); liquid water versus ice crystal \citep{Cheng2019};
and \ce{Mn2ScSe} vs \ce{CrPt2Sc} structures taken from the WBM dataset \citep{Wang2021}.
For each system, cells of the smallest to the largest sizes are built by supercell replication along with all of their atoms inside,
and test structures are sampled from MD driven by a foundation potential \citep{Takamoto2022,Rhodes2025}.
The models are trained on the ID structures.
Dataset generation and model training are described in Appendix~\ref{app:datasetgen} \& \ref{app:mliptraining}.

\begin{figure}[t]
  \centering
  \includegraphics[width=\linewidth]{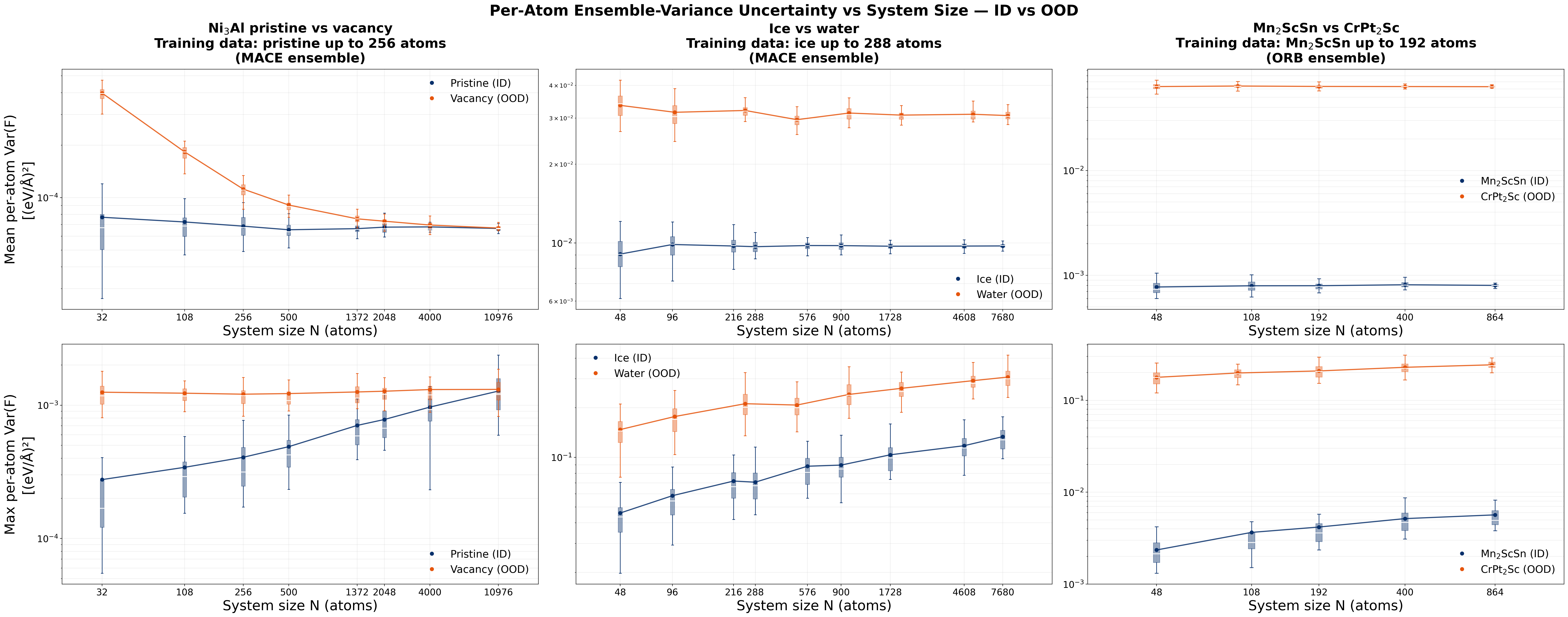}
  \caption{
  System size introduces bias to uncertainty, eliminating the ID/OOD gap and failing size-unaware-UQ-based OOD detection at large sizes.
  Per-atom uncertainty versus system size $N$ for three ID/OOD pairs:
  pristine versus vacancy Ni$_3$Al at 500K (left, MACE ensemble variance), liquid water
  versus ice (middle, MACE ensemble variance), and a \ce{Mn2ScSe} structure
  versus a \ce{CrPt2Sc} structure (right, ORB ensemble variance). Each box plot
  summarizes 40 test structures sampled from MD at that size. \textbf{Top:} mean uncertainty $\bar{u}$, flat for
  homogeneous systems and decaying towards the pristine value for the vacancy
  system. \textbf{Bottom:} maximum uncertainty $u_{\max}$, increasing with system size $N$ for
  every homogeneous ID system while remaining flat for the vacancy system.
  }
  \label{fig:scaling}
  \vspace{-4mm}
\end{figure}

The top row of Figure~\ref{fig:scaling} shows $\bar{u}$ against $N$. For all
homogeneous systems, $\bar{u}$ remains almost constant against system size $N$.
Replicating a homogeneous cell resamples the same environment distribution.
The vacancy system instead starts high in small cells and decays towards the pristine
value.
A fixed number of defect-perturbed atoms contributes to the mean as
$\mathcal{O}(1/N)$, and in small supercells the vacancy interacts with its
periodic images, distorting a larger fraction of the cell --- the finite-size
effect that motivates image and elastic corrections in first-principles
defect calculations \citep{Leslie1985,Freysoldt2014}.
In large systems, mean uncertainty $\bar{u}$ can no longer distinguish a defective structure from a pristine, hence requiring a statistic sensitive to individual atoms, such as $u_{\max}$ to detect a local error.
 
The bottom row of Figure~\ref{fig:scaling} shows the opposite trend for $u_{\max}$. For every
homogeneous ID system, $u_{\max}$ increases with $N$, while the vacancy system stays flat.
The gap between small OODs and large IDs disappears as the ID size grows, failing OOD detection. 
This shows a practical limitation of current size-unaware UQ methods in OOD detection.
Notably, the distributions do not overlap within the sizes in our calculations for the WBM pair, but given the present scaling, crossing between small OODs and large IDs is in principle reachable at large enough ID sizes in large-scale MLIP-driven MD.
Alternatively, percentile-based UQs show mean UQ behavior at smaller sizes and max UQ behavior at larger sizes, hence are still not capable of OOD detection at large sizes (Appendix \ref{app:pc-uq}).

We suggest that \textbf{extreme-value statistics} may explain this trend.
$u_{\max}$ is the largest of $N$ draws from the per-atom uncertainty
distribution, and for a broad class of distributions the expected maximum
grows slowly with $N$ \citep{Fisher1928,Coles2001}: a larger
system reaches further into the tail of its own environment distribution,
surfacing rarer and more distorted local environments. A defect, diluted with
system size, instead adds a fixed number of distorted environments whose
uncertainty does not change with cell size, matching the flat $u_{\max}$ we
observe.

To decide whether the rising $u_{\max}$ marks local atomic environment that are truly unfamiliar to
the model, we track how the local environment and
true force error of the max-UQ atom change with system size and find both grow
(Appendix~\ref{app:desc-and-error}).
This means the atom moves further from the training
set in SOAP descriptor space \citep{Bartok2013} and its force error rises well
above those of smaller sizes.
\textbf{The high uncertainty is therefore valid}, and any
uncertainty-thresholded selection rule
\citep{Podryabinkin2017,Smith2018,Zhang2019,Vandermause2020,Kulichenko2023}
would flag these structures for ground-truth calculation for model retraining,
which we test in Section~\ref{sec:training}.

\vspace{-1mm}
\subsection{Training Structure Size Effect}
\label{sec:training}
\vspace{-1mm}

Section~\ref{sec:scaling} showed that the max-UQ atoms of large in-distribution (ID)\footnote{We will keep calling these structures ID because the structure as a whole is still the same phase as the smaller structures in the training set.
Strictly speaking, the max-UQ atom itself is OOD: its local environment and force error
are distinct from what the model is trained on.} systems have unfamiliar local environment and high force errors, thus a standard active learning loop would send them for ground-truth calculation.
We test this by retraining the ORB ensemble on Ni$_3$Al under three settings and recomputing $u_{\max}$ across sizes: a baseline trained on pristine systems up to 256 atoms, the same baseline plus 500-atom pristine systems, and the baseline plus vacancy systems up to 256 atoms. The first two differ only in whether the large ID
structures the uncertainty flagged are labeled as OOD for retraining; the third trains the models on small structures that are OOD by construction.

\begin{figure}[t]
  \centering
  \includegraphics[width=\linewidth]{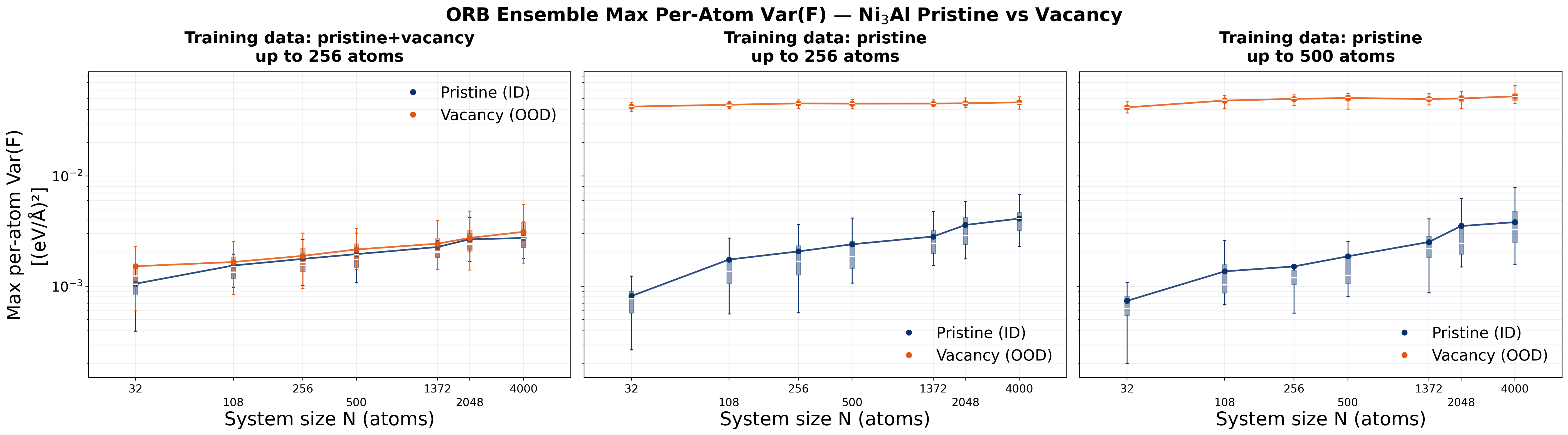}
  \caption{Maximum per-atom force variance of a ORB ensemble for pristine and
  vacancy Ni$_3$Al against system size, under three training sets. Adding
  500-atom pristine systems (right) leaves the pristine curve close to the
  baseline (middle), only reducing UQ by
  $\approx10^{-4}~(\mathrm{eV/\text{\normalfont\AA}})^2$ from, while adding vacancy systems up to 256 atoms (left) significantly lowers
  the vacancy uncertainty by about two orders of magnitude larger at $\approx10^{-2}~(\mathrm{eV/\text{\normalfont\AA}})^2$, closing most of the gap to the pristine curve.
    }
  \label{fig:training}
  \vspace{-4mm}
\end{figure}

As shown in Figure \ref{fig:training}, adding the 500-atom pristine systems leaves the pristine $u_{\max}$ at 500 atoms at roughly the same value as the baseline, and the scaling with $N$ still exists.
This holds even though the model has now seen these structures directly, and even though their max-UQ atoms were shown in Section~\ref{sec:scaling} to sit far from the training set with
force errors above smaller sizes.
Training on smaller vacancy systems instead lowers the
vacancy $u_{\max}$ significantly at every
size by about two orders of magnitude larger than training on large IDs, enough to close most of the gap to the pristine curve at 500 atoms.
For this system and model, \textbf{labeling and training on a small defect-type OOD are computationally cheaper and improve the models much more significantly than training on a size-inflated ID}.
The same pattern also holds for other systems and models
(Appendix~\ref{app:training-size-effects}).

\textbf{The mean squared error loss} used to train the
models, which averages over all atoms, could be a reason
\citep{Behler2007,Batatia2022}.
A handful of distorted environments in a 500-atom system contributes $\mathcal{O}(1/N)$ to the total loss, so the optimizer can leave them poorly fit while the mean error stays low,
and adding more such structures need not change that balance.
This is consistent with imbalanced regression more broadly, where rare targets are sacrificed for bulk
accuracy unless the loss is reweighted \citep{Lin2017,Yang2021}.
Nevertheless, given the current widely-adopted mean-based training loss, our results show a \textbf{limitation of current UQ methods that do not take size into account for OOD detection}.

\textbf{Two practical guidelines thus follow for active learning}.
First, large ID structures whose $u_{\max}$ is inflated by size appear to be poor candidates for ground-truth calculation under the standard training setup:
they are flagged by uncertainty, have unseen local atomic environment and high error, and still yield
little model improvement.
Second, addressing this requires either removing the size dependence from the uncertainty so that thresholds compare like to like by means of, e.g., post-hoc trainable uncertainty calibrator,
or changing the loss so that outlier max-UQ atoms are fit without sacrificing bulk accuracy.
We implement a \textbf{trainable post-hoc uncertainty calibrator} using atomic descriptors for system detection (Appendix \ref{app:calibrator}) and find it not trivial to remove the size-induced bias on uncertainty values correctly for all systems, remaining an area for future studies.

\vspace{-3mm}
\section{Conclusions}
\label{sec:conclusions}
\vspace{-1mm}
We investigate how per-atom force uncertainty scales with system size across different datasets and MLIPs.
Mean uncertainty remains constant for homogeneous systems and dilutes with size for defective ones, thus cannot detect a local failure mode in the large systems.
Maximum uncertainty is sensitive to local error but carries size dependence: it increases with $N$ for ID systems while staying flat for a defect. This size effect eliminates the ID/OOD gap at large sizes and could lead to \textbf{incorrect uncertainty-based OOD detection}.
We suggest \textbf{extreme-value statistics} as an explanation, since a larger system samples deeper into the tail of its own environment distribution.

The max-UQ atoms are actually valid as OOD candidates, considering both errors and atomic environments, for ground-truth calculation and model retraining, which we nonetheless find to barely improve the model.
In contrast, retraining on small OOD-by-design structures is cheaper and improves the model more substantially.
We attribute this to the mean-based loss used to train MLIPs,
which tolerates a few poorly fit atoms as long as the average error stays low.
This exposes \textbf{a limitation of current size-unaware UQ methods in OOD detection}, given the current training.

We subsequently provide a practical guideline for active learning on mixed-size datasets that \textbf{size-unaware $u_{\max}$ should not be compared across sizes with a single threshold}.
Two immediate solutions are possible: removing the size bias from the uncertainty, e.g., with a post-hoc size-aware calibrator which we find to be not trivial to implement,
or devising a training loss that can fit outlier atoms without sacrificing bulk accuracy, both of which we leave for future work to improve OOD detection in large-scale MLIP-accelerated materials simulations and discovery.

\bibliographystyle{unsrt}
\bibliography{references}


\appendix

\section{Dataset Generation}
\label{app:datasetgen}

For each system, a starting unit cell is expanded into supercells of increasing size by replication, and NPT molecular dynamics (MD) simulation is run independently at each size for 20,000 steps, saving one structure every 100 steps, for 200 structures per size.
Each set of 200 structures is randomly split into 120 for training, 40 for validation, and 40 for testing.
The same total number of MD steps, snapshot frequency, and split sizes are used across all systems,
so that Section~\ref{sec:scaling} compares uncertainty across sizes without any difference in how much data went into training or testing at each size.

Pristine, vacancy, and cubic Ni$_3$Al structures start from the mp-2593 conventional standard cell \citep{Jain2013,Mishin2004} and are expanded into larger supercells.
MD is run with a Matlantis PFP v8.0.0 R2SCAN ASE calculator \citep{Takamoto2022} at 300~K and with a timestep of 2~fs.
Vacancy structures are generated by
removing one atom from each supercell prior to MD.

Liquid water and ice structures are generated the same way, using an ORB v3-direct-20-mpa calculator \citep{Rhodes2025} in place of PFP.
The timestep for MD at 300K is reduced to 0.5~fs because of the light hydrogen atoms; all other
settings (total steps and snapshot frequency) match the
Ni$_3$Al runs except ice structures that are run at 250~K.

The WBM structures \citep{Wang2021,Riebesell2025}
are each expanded into supercells and run with the same ORB v3-direct-20-mpa calculator and the same timestep as the Ni$_3$Al runs.

For all systems, energies and forces used as training and test labels are the calculator predictions above (PFP for Ni$_3$Al, ORB v3-direct-20-mpa for water/ice and WBM), rather than DFT.
This keeps labeling tractable across the full range of system sizes and lets us compute the true force error of every atom, at every size, against a fixed reference.

\section{MLIP Training}
\label{app:mliptraining}

For each system and size setting, we train two force ensembles, MACE and ORB (direct forces), with five independently seeded initial members each, plus a single ORB hybrid model used for the Force-Delta uncertainty estimate.
All members within an ensemble share the same architecture and optimizer settings and differ only in random seed (weight initialization and data shuffling). 
Training and validation use the 120 and 40 structures set aside in Appendix~\ref{app:datasetgen};
the remaining 40 held-out test structures are used for the size-scaling analysis in Sections~\ref{sec:scaling} and \ref{sec:training} and are never seen during training or model selection.

MACE ensemble members are ScaleShiftMACE models with two interaction layers, 128 channels, and an $L_{\max}=1$, correlation-3 equivariant expansion, following the standard MACE-medium architecture \citep{Batatia2022}.
Isolated-atom energies (E0s) are set by averaging over the training set rather than fit by linear regression.
Both the MACE and ORB ensembles predict energies and
forces from a single output;
the ORB hybrid model additionally predicts a second, energy-gradient force so that the direct and conservative forces of the same model are trained and can be compared \citep{Lortaraprasert2025}.
All models are trained with AdamW \citep{Loshchilov2019} using a weighted sum of per-atom energy and force losses, gradient clipping, and early stopping on the validation loss.
Architecture and training settings are summarized in
Table~\ref{tab:training_hparams}.
All trainings converge smoothly.

\begin{table}[t]
  \centering
  \caption{Architecture and training settings for the MACE ensemble, the ORB ensemble (direct forces only), and the ORB hybrid model (both direct and conservative forces trained, used for the Force Delta uncertainty estimate).}
  \hspace*{-1.7cm}
  \label{tab:training_hparams}
  \begin{tabular}{lccc}
    \toprule
     & MACE ensemble & ORB ensemble & ORB hybrid \\
    \midrule
    Backbone & ScaleShiftMACE & orb-v3-direct (from scratch) & orb-v3-direct (from scratch) \\
    Cutoff radius (\AA) & 6.0 & 6.0 & 6.0 \\
    Radial basis & Bessel (10) & Bessel (8) & Bessel (8) \\
    Angular basis & $\ell_{\max}=3$ & spherical harmonics, $\ell_{\max}=3$ & spherical harmonics, $\ell_{\max}=3$ \\
    Channels / latent dim & 128 & 256 & 256 \\
    Interaction / MP layers & 2 & 5 & 5 \\
    Activation & SiLU & SiLU & SiLU \\
    Energy loss & MSE & Huber ($\delta=0.01$) & Huber ($\delta=0.01$) \\
    Force loss & MSE & conditional Huber ($\delta=0.01$) & conditional Huber ($\delta=0.01$) \\
    Conservative-force loss weight & --- & 0.0 & 1.0 \\
    Optimizer & AdamW & AdamW & AdamW \\
    Learning rate & $10^{-4}$ & $10^{-4}$ & $10^{-4}$ \\
    Weight decay & $10^{-5}$ & $10^{-5}$ & $10^{-5}$ \\
    Batch size & 1 & 16 (effective) & 16 (effective) \\
    Gradient clip & 10.0 & 10.0 & 10.0 \\
    Max epochs & 50 & 50 & 50 \\
    Early-stopping patience & 20 & 20 & 20 \\
    Ensemble size & 5 & 5 & 1 \\
    \bottomrule
  \end{tabular}
\end{table}

\section{Additional Results}
\subsection{Atomic Descriptors and Force Errors of Max-Uncertainty Atoms}
\label{app:desc-and-error}

For each test structure, we take the atom with the largest ensemble force
variance, the same atom that sets $u_{\max}$ in Section~\ref{sec:scaling}, and
examine its force error, and local environment.
Force error is the absolute error between the ensemble mean force and the reference label from Appendix~\ref{app:datasetgen}. Environment
novelty is measured with a SOAP descriptor \citep{Bartok2013}, projected onto
principal components fit on the training set; we report the magnitude of the
projection onto the second component (PC2), which explains 37.2\% of the
variance in the training descriptors and separates bulk from distorted
environments most clearly.

\begin{figure}[t]
  \centering
  \includegraphics[width=\linewidth]{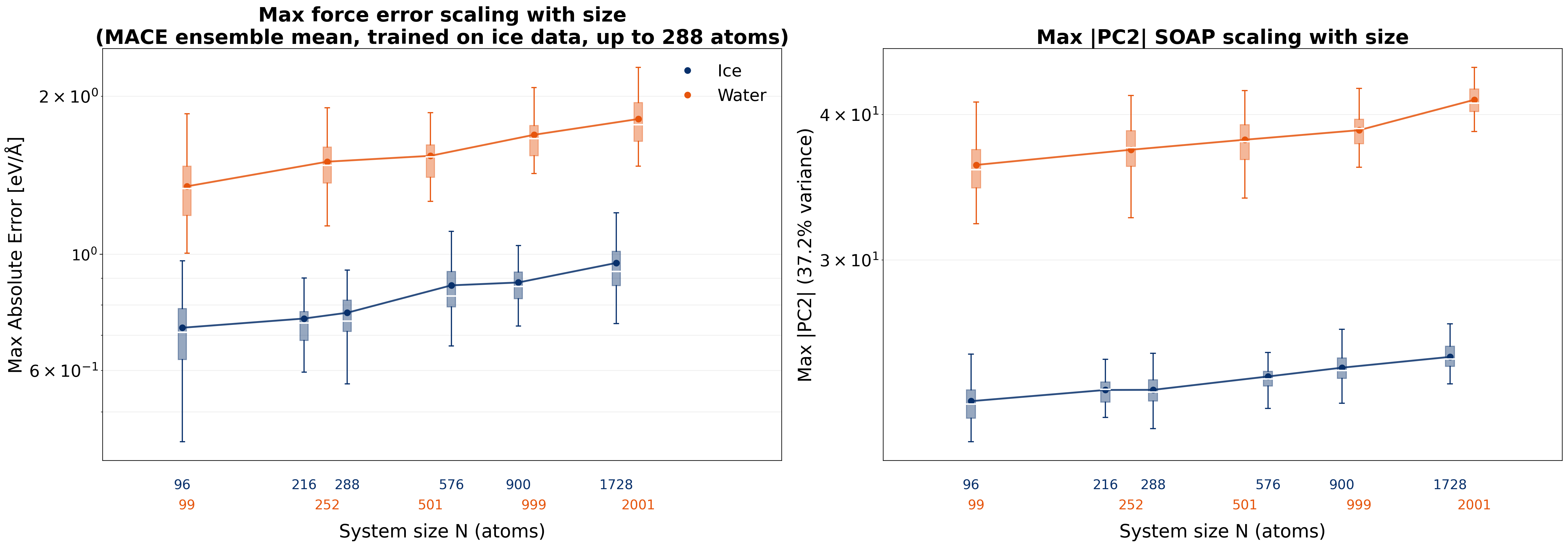}
  \caption{Left: maximum absolute force error of the MACE ensemble mean against
  system size, for the atom with the largest ensemble variance in each test
  structure, for ice (ID, trained up to 288 atoms) and water (OOD). Right:
  maximum absolute value of the same atom's SOAP PC2 projection, against system
  size. Both quantities increase with $N$ for both systems. Boxes summarize the
  same test structures as Figure~\ref{fig:scaling}.}
  \label{fig:tail_ice_water}
\end{figure}

Figure~\ref{fig:tail_ice_water} shows both quantities against system size for
the ice/water pair, where the MACE ensemble is trained on ice up to 288 atoms
and tested up to 1728 atoms, with water never seen in training. Both the force
error and the PC2 magnitude of the max-UQ atom increase with $N$ for both
systems. This holds for water, which is out of distribution by phase, and for
ice, which is in distribution by phase but is tested well beyond its training
size range. In both cases, the atom flagged as most uncertain is also the atom
that is furthest from the training descriptors and worst predicted, which shows the increase in
$u_{\max}$ tracks genuine, not merely statistical, novelty as OOD candidates.

\subsection{Force Delta Scaling with System Size}
\label{app:force-delta}
The ORB hybrid force-delta metric requires no ensemble, with the distance between its direct and energy-gradient forces taken directly as the uncertainty estimate.
For each ID/OOD pair, the hybrid model is trained  on ID data only, as in Section~\ref{sec:scaling}.

\begin{figure}[t]
  \centering
  \includegraphics[width=\linewidth]{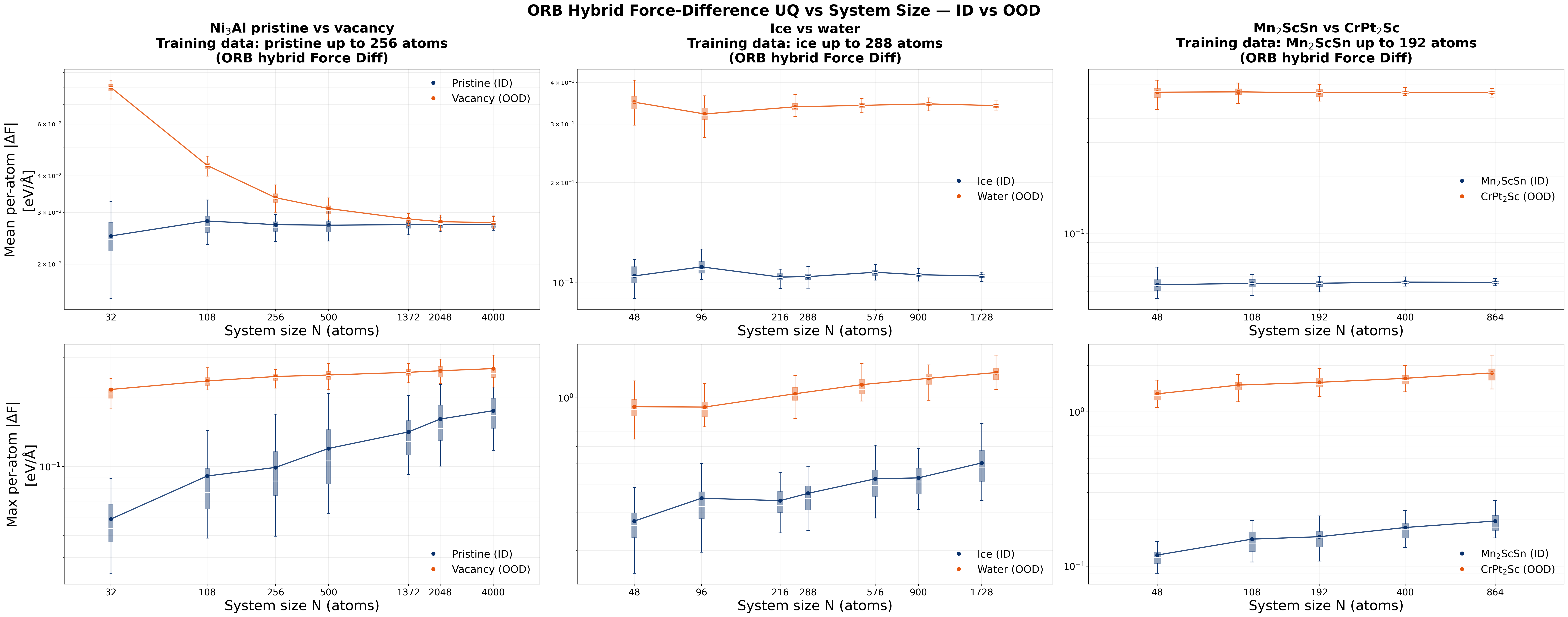}
  \caption{Mean (top) and maximum (bottom) per-atom ORB force delta against
  system size, for three ID/OOD pairs: pristine versus vacancy Ni$_3$Al (left,
  trained on pristine up to 256 atoms), ice versus water (middle, trained on ice up to 288 atoms), and Mn$_2$ScSn versus CrPt$_2$Sc (right, trained
  on Mn$_2$ScSn up to 192 atoms). Boxes summarize the same test structures as
  the corresponding ensemble-variance results in Section~\ref{sec:scaling}.}
  \label{fig:force_delta_appendix}
\end{figure}

The top row of Figure~\ref{fig:force_delta_appendix} shows the mean force delta, and the bottom row the maximum, against system size, for all three ID/OOD pairs. The pattern matches the ensemble variance results in Section~\ref{sec:scaling}. Mean force delta for the Ni$_3$Al vacancy dilutes toward the pristine value as $N$ grows, consistent with a fixed number of defect atoms contributing less to the mean as the cell size increases. Mean force delta is flat and well separated between ID and OOD for the two homogeneous pairs, ice/water and Mn$_2$ScSn/CrPt$_2$Sc, since there is no dilution effect when the whole structure is out of distribution. Maximum force delta rises with $N$ for every ID system across all three pairs, while the vacancy OOD stays roughly flat, again narrowing the ID--OOD gap from below; for Ni$_3$Al the ID and OOD boxes overlap at the largest sizes tested. That this same scaling appears for a metric that needs no ensemble, indicates the effect is not specific to ensemble variance.

\subsection{Percentile-Based Uncertainty}
\label{app:pc-uq}
\begin{figure}[t]
  \centering
  \includegraphics[width=\linewidth]{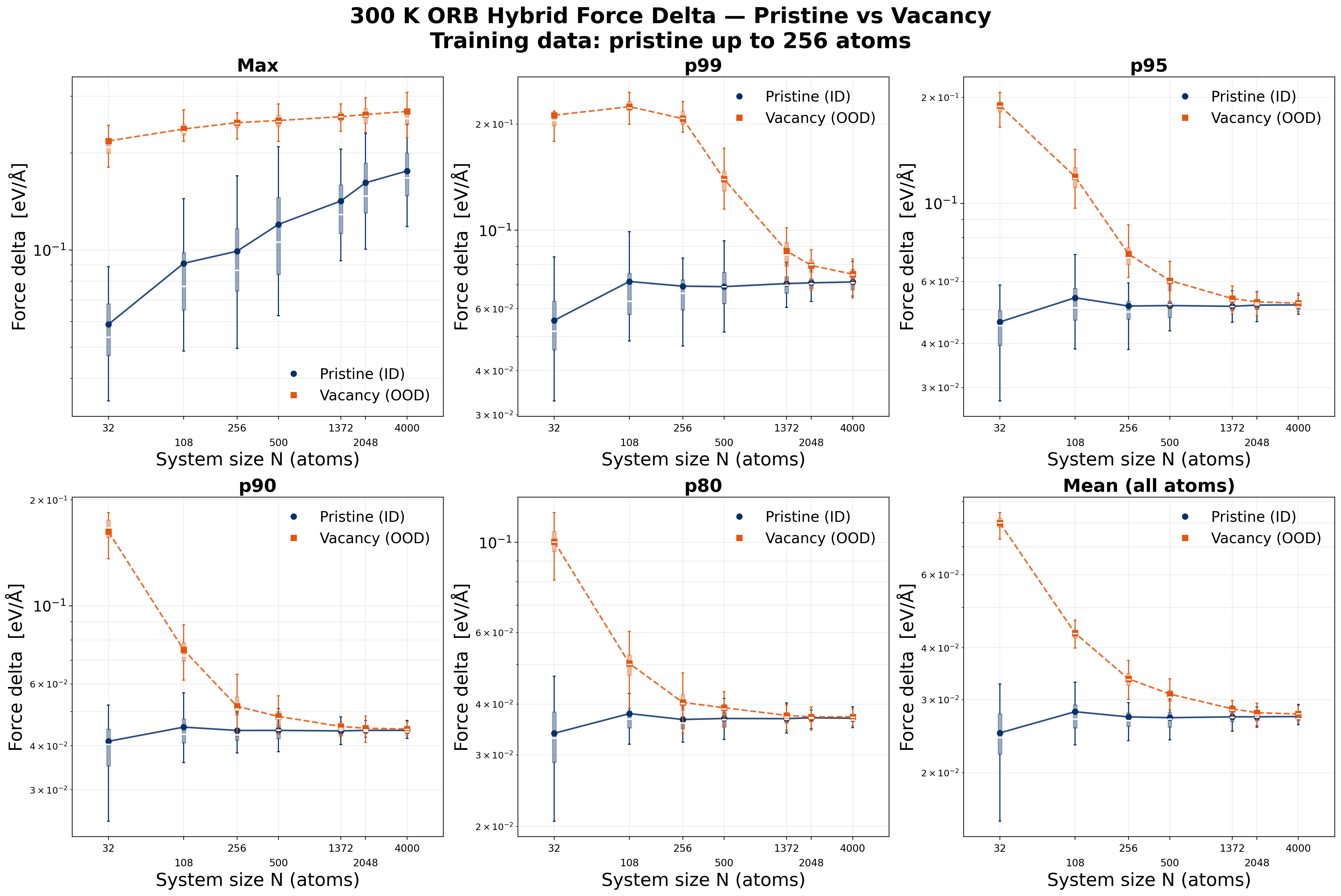}
  \caption{Per-atom ORB force delta for pristine and vacancy Ni$_3$Al against
  system size, pooled at different percentiles (max, p99, p95, p90, p80, and
  mean). Lower percentiles converge to the pristine value at smaller system
  sizes, resembling the dilution seen for the mean in
  Section~\ref{sec:scaling}, while the max resists dilution and remains
  separated from the vacancy curve throughout.}
  \label{fig:percentile_appendix}
\end{figure}

Figure~\ref{fig:percentile_appendix} extends the mean/max comparison in
Section~\ref{sec:scaling} to intermediate percentiles (p80, p90, p95, p99) of
the per-atom force delta distribution, for the same pristine/vacancy Ni$_3$Al
pair. At small system sizes, every percentile shown, including p80, separates
vacancy from pristine as clearly as the max does: the defect sets a large
fraction of the distribution, so any high percentile falls on a
defect-perturbed atom. As $N$ grows, this separation collapses at different
rates depending on the percentile. p80 and p90 converge to the pristine curve
almost immediately, by 256 atoms; p95 and p99 persist further but still
converge by around 1000--2000 atoms. Only the true max keeps rising with $N$
for the pristine system and stays mostly flat for the vacancy, so the ID--OOD gap
narrows but never closes to zero within the sizes tested.

The reason is that a percentile, unlike the max, is diluted by system size in
the same way the mean is: a fixed number of defect-perturbed atoms occupies a
shrinking fraction of the ranked distribution as $N$ grows, so eventually the
defect atoms fall below the percentile cutoff and the score reverts to the
bulk value. The max is immune to this because it always tracks the single
worst atom, regardless of how many atoms are averaged into the rank below it.
Percentile-based pooling therefore behaves like the max at small system sizes,
where it still detects the defect, but like the mean at large system sizes,
where it dilutes and loses the defect signal, with the crossover size set by
how extreme the percentile is.

\subsection{Training Structure Size Effect on Other Systems and Models}
\label{app:training-size-effects}
The same three training settings used for Ni$_3$Al in Section~\ref{sec:training} are repeated here for two other UQ estimators, the ORB ensemble force variance and the ORB hybrid force delta, and for two other ID/OOD pairs: ice versus water (MACE ensemble) and Mn$_2$ScSn versus CrPt$_2$Sc, the two WBM structures introduced in Section~\ref{sec:scaling} (ORB ensemble and hybrid). The pattern is the same in every case: adding larger ID structures to the training set has almost no effect on either curve, while adding OOD-type structures reduces OOD uncertainty and narrows the gap to the ID curve.

\begin{figure}[t]
  \centering
  \includegraphics[width=\linewidth]{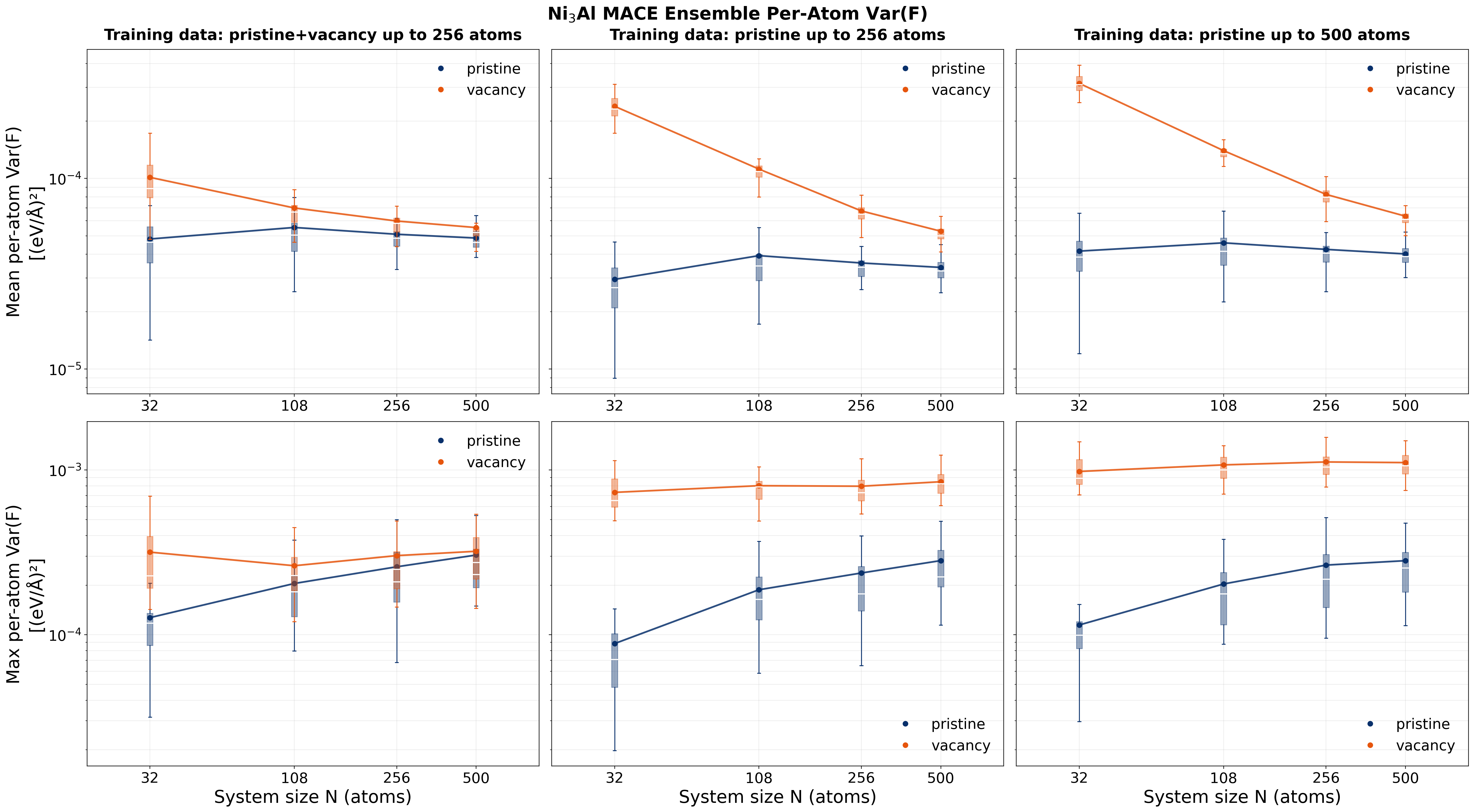}
  \caption{MACE ensemble force variance, mean (top) and maximum (bottom)
  per-atom, for pristine (ID) versus vacancy (OOD) Ni$_3$Al, under the same
  three training settings as Figure~\ref{fig:training}: pristine+vacancy up to
  256 atoms (left), pristine-only up to 256 atoms (middle), and pristine-only
  up to 500 atoms (right).}
  \label{fig:app_niAl_mace_ens}
\end{figure}

\begin{figure}[t]
  \centering
  \includegraphics[width=\linewidth]{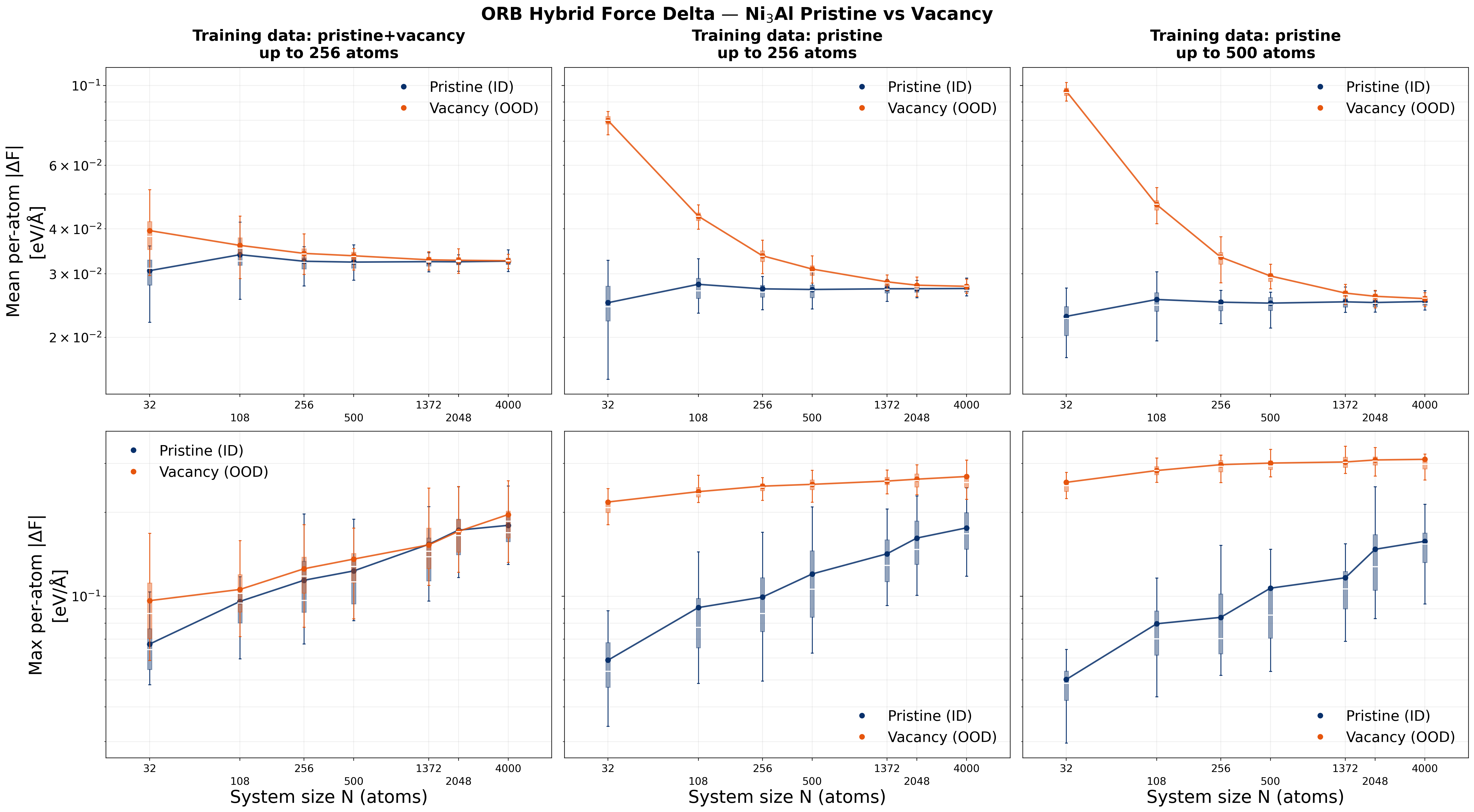}
  \caption{ORB hybrid force delta, mean (top) and maximum (bottom) per-atom,
  for pristine (ID) versus vacancy (OOD) Ni$_3$Al, under the same three
  training settings as Figure~\ref{fig:training}.}
  \label{fig:app_niAl_orb_hybrid}
\end{figure}

For pristine/vacancy Ni$_3$Al, this holds for both the MACE ensemble variance (Figure~\ref{fig:app_niAl_mace_ens}) and the ORB hybrid force delta (Figure~\ref{fig:app_niAl_orb_hybrid}): training on pristine up to 500 atoms instead of 256 leaves both curves close to unchanged, while training on pristine plus vacancy nearly closes the gap between them, matching the ORB result in Figure~\ref{fig:training}.

\begin{figure}[t]
  \centering
  \includegraphics[width=\linewidth]{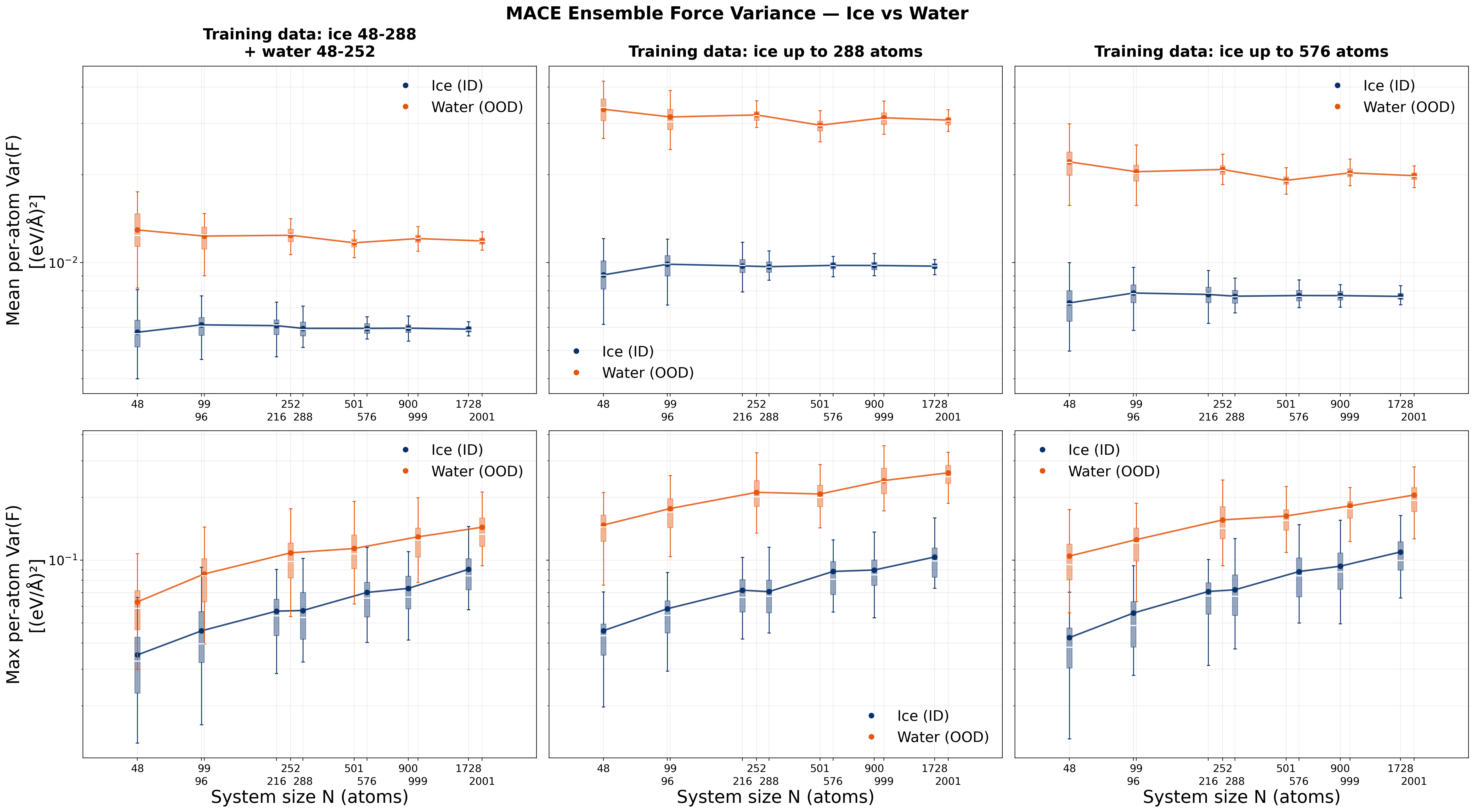}
  \caption{MACE ensemble force variance, mean (top) and maximum (bottom)
  per-atom, for ice (ID) versus water (OOD), under three training settings:
  combined ice (48--288 atoms) and water (48--252 atoms) (left), ice-only up to
  288 atoms (middle, same setting as Figure~\ref{fig:tail_ice_water}), and
  ice-only up to 576 atoms (right).}
  \label{fig:app_icewater}
\end{figure}

For ice versus water (Figure~\ref{fig:app_icewater}), training on ice up to 576 atoms instead of 288 leaves both curves close to unchanged, with a gap similar to Figure~\ref{fig:tail_ice_water}. Training on a combined ice+water dataset narrows the gap but, unlike the vacancy case, does not close it: water uncertainty stays above ice at every size.

\begin{figure}[t]
  \centering
  \includegraphics[width=\linewidth]{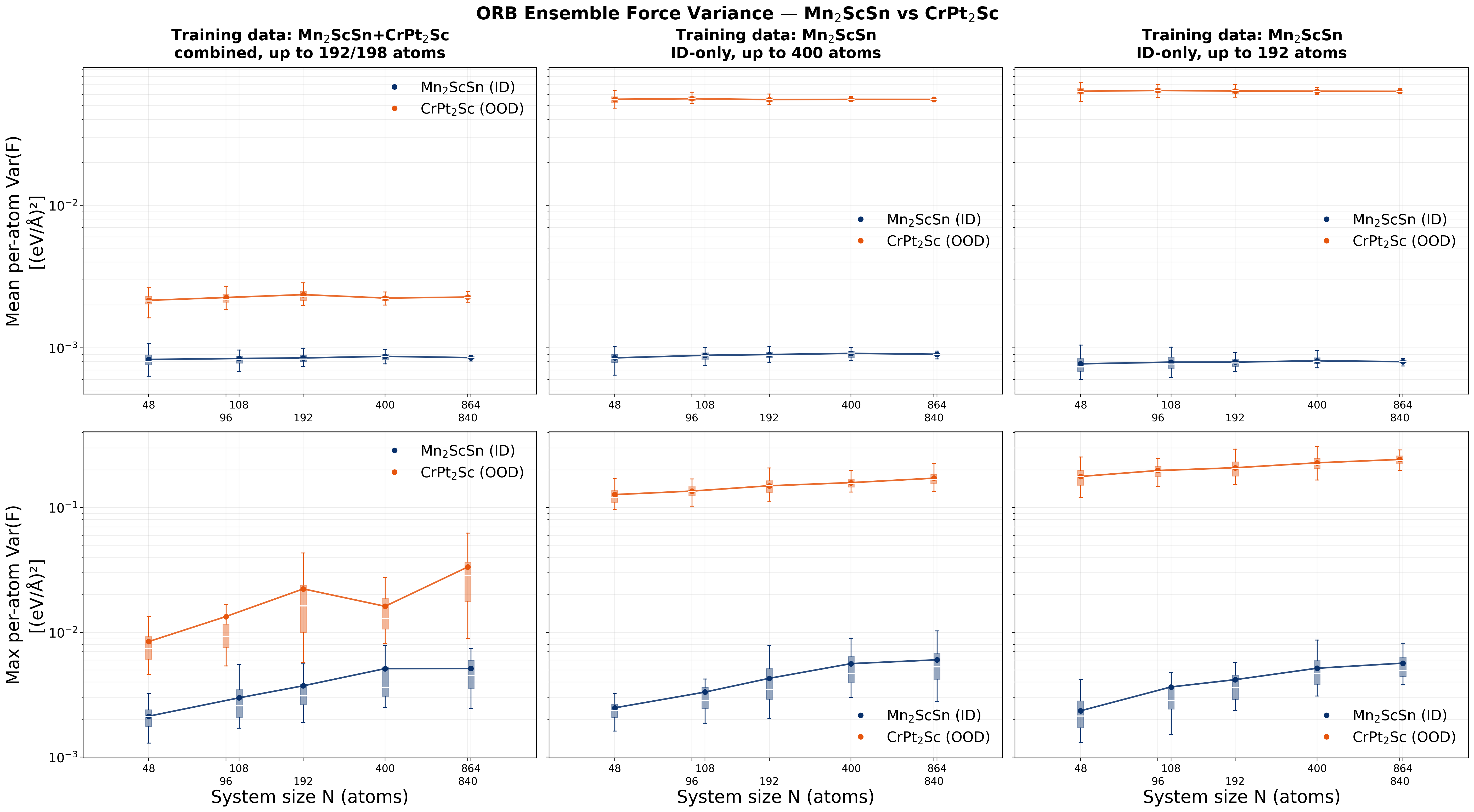}
  \caption{ORB ensemble force variance, mean (top) and maximum (bottom)
  per-atom, for Mn$_2$ScSn (ID) versus CrPt$_2$Sc (OOD), under three training
  settings: combined training data up to 192/198 atoms (left), Mn$_2$ScSn-only
  up to 400 atoms (middle), and Mn$_2$ScSn-only up to 192 atoms (right).}
  \label{fig:app_mn2scsn_orb_ens}
\end{figure}

\begin{figure}[t]
  \centering
  \includegraphics[width=\linewidth]{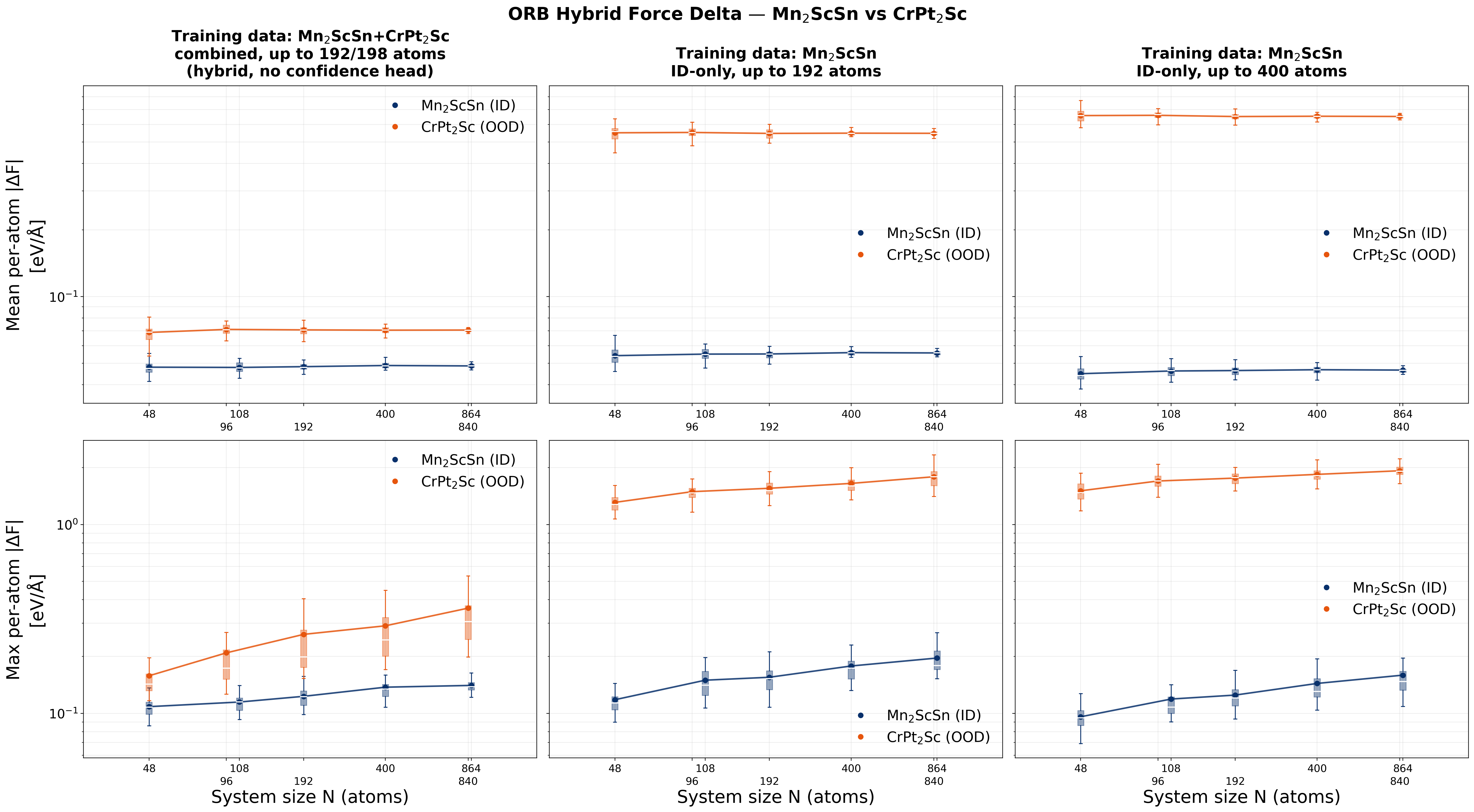}
  \caption{ORB hybrid force delta, mean (top) and maximum (bottom) per-atom,
  for Mn$_2$ScSn (ID) versus CrPt$_2$Sc (OOD), under three training settings:
  combined training data up to 192/198 atoms (left), Mn$_2$ScSn-only up to 192
  atoms (middle), and Mn$_2$ScSn-only up to 400 atoms (right).}
  \label{fig:app_mn2scsn_orb_hybrid}
\end{figure}

For Mn$_2$ScSn versus CrPt$_2$Sc (Figures~\ref{fig:app_mn2scsn_orb_ens} and~\ref{fig:app_mn2scsn_orb_hybrid}), ID-only training at either size leaves close to an order-of-magnitude gap between the two structures, for both UQ metrics. Combined training reduces this gap substantially but again does not fully close it.

Across three ID/OOD pairs, two model architectures, and two UQ estimators, the same qualitative pattern holds: training on more of the same type of structure does not help, and training on structures unlike the training set does, though how completely it closes the gap depends on the system.

\subsection{Trainable Post-Hoc Uncertainty Calibrator}
\label{app:calibrator}

As a preliminary test of whether the size dependence in Section~\ref{sec:scaling}
can be removed post hoc, we implement a trainable calibrator that takes an
atom's local environment and system size as input and outputs a corrected
uncertainty. The environment is summarized by projecting its SOAP descriptor
\citep{Bartok2013} onto its leading PCA components, giving a low-dimensional
feature $x = \{x_{\text{mean}}; x_{\text{argmaxUQ}}\}$; system size enters as $N$. All uncertainty values, $N$, and the
calibrator outputs are worked with in log scale throughout.

The calibrator has two parts. An epsilon neural network $\text{NN}_{\text{eps}}$ is
trained to predict the raw log-uncertainty directly from $(x, N)$, fitting the
size-dependent trend itself: for every observed $(x, N, U_{\text{orig}})$
triple, its loss drives $\text{NN}_{\text{eps}}(x, N)$ toward
$U_{\text{orig}}$. A corrector network $\text{NN}_{\text{cor}}$ then outputs an
additive correction, $U_{\text{cal}} = U_{\text{orig}} + \text{NN}_{\text{cor}}(x, N)$,
trained so that $U_{\text{cal}}$ matches what the epsilon network would predict
for the same environment at a fixed reference size, $\text{NN}_{\text{eps}}(x, N{=}100)$.
In effect, the corrector learns to shift an atom's uncertainty to what it would
have been in a reference-size structure with the same local environment,
removing the size trend rather than the environment signal.

\begin{figure}[t]
  \centering
  \includegraphics[width=\linewidth]{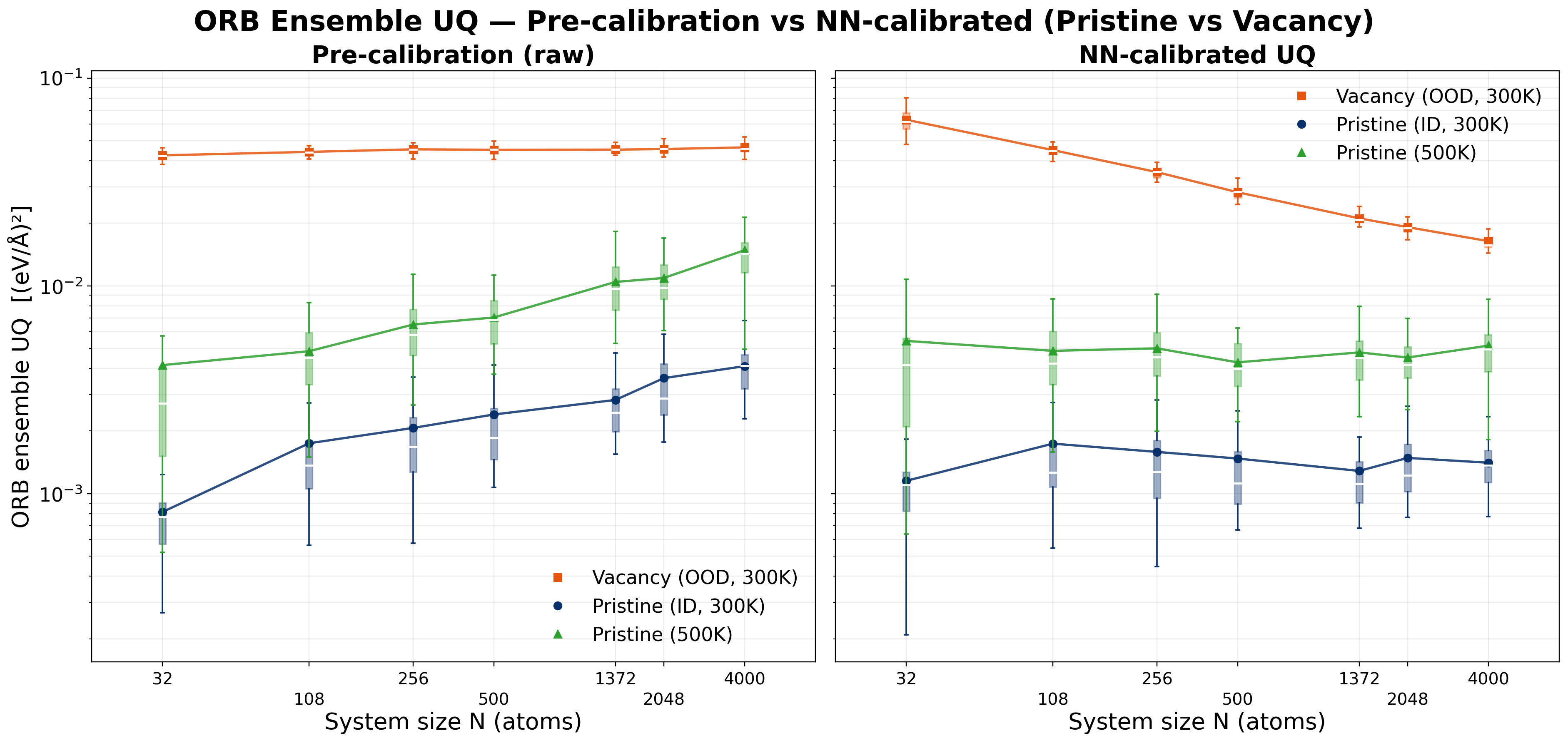}
  \caption{ORB ensemble uncertainty for pristine Ni$_3$Al at 300~K (ID) and
  500~K, and vacancy Ni$_3$Al at 300~K (OOD), before (left) and after (right)
  size-aware calibration. Calibration flattens the pristine curves across
  size but also lowers the vacancy curve, narrowing the ID/OOD gap as $N$
  grows.}
  \label{fig:calibrator}
\end{figure}

We test this on the ORB ensemble trained on pristine Ni$_3$Al at 300~K
(Section~\ref{sec:training}), training the calibrator on raw uncertainties from
both 300~K and 500~K pristine structures across all sizes. Figure~\ref{fig:calibrator}
compares the raw and calibrated uncertainty for pristine and vacancy
structures. Calibration removes most of the size trend for the pristine
systems at both temperatures, flattening curves that otherwise grow by an
order of magnitude across our size range. However, the same correction also
pulls down the vacancy curve, which decreases with $N$ instead of staying
flat, and the ID/OOD gap that was roughly constant before calibration
narrows steadily with size afterward. Since the calibrator has no explicit
notion of OOD, it corrects the vacancy curve as if it followed the same
size trend as pristine data, which is not the desired behavior. This is a
proof of concept rather than a solution: it shows that size-aware calibration
can flatten the pristine trend, but also that doing so without distorting
genuine OOD signal is not trivial, motivating further work on calibrators that
separate the two.

\end{document}